\documentclass{article}
\usepackage{ijcai25}

\usepackage{times}
\usepackage{soul}
\usepackage{url}
\usepackage[hidelinks]{hyperref}
\usepackage[utf8]{inputenc}
\usepackage[small]{caption}
\usepackage{graphicx}
\usepackage{amsmath}
\usepackage{amsthm}
\usepackage{booktabs}
\usepackage{algorithm}
\usepackage{algorithmic}
\usepackage[switch]{lineno}

\usepackage{subfigure}
\usepackage{multirow}
\usepackage{amssymb}
\usepackage{xcolor}
\usepackage{overpic}
\usepackage{makecell}
\makeatletter
\renewcommand{\maketag@@@}[1]{\hbox{\m@th\normalsize\normalfont#1}}%
\makeatother

\title{Base-Detail Feature Learning Framework for Visible-Infrared Person Re-Identification}

\author{
Zhihao Gong$^1$
\and
Lian Wu$^2$\and
Yong Xu$^{3,*}$\\
\affiliations
$^1$Harbin Institute of Technology (Shenzhen)\\
$^2$GuiZhou Education University\\
$^3$Harbin Institute of Technology (Shenzhen)\\
\emails
gongzhh888@gmail.com, 
wulian\_best@163.com, 
laterfall@hit.edu.cn
}

\begin{document}

\maketitle

\begin{abstract}
Visible-infrared person re-identification (VIReID) provides a solution for ReID tasks in 24-hour scenarios; however, significant challenges persist in achieving satisfactory performance due to the substantial discrepancies between visible (VIS) and infrared (IR) modalities. Existing methods inadequately leverage information from different modalities, primarily focusing on digging distinguishing features from modality-shared information while neglecting modality-specific details. To fully utilize differentiated minutiae, we propose a Base-Detail Feature Learning Framework (BDLF) that enhances the learning of both base and detail knowledge, thereby capitalizing on both modality-shared and modality-specific information. Specifically, the proposed BDLF mines detail and base features through a lossless detail feature extraction module and a complementary base embedding generation mechanism, respectively, supported by a novel correlation restriction method that ensures the features gained by BDLF enrich both detail and base knowledge across VIS and IR features. Comprehensive experiments conducted on the SYSU-MM01, RegDB, and LLCM datasets validate the effectiveness of BDLF.
\end{abstract}

\section{Introduction}
Person re-identification (ReID) aims to retrieve a target identity from gallery images captured by different cameras \cite{b2} and has recently demonstrated significant advancements in the fields of security and public surveillance \cite{b3}. However, most existing methods \cite{b1}\cite{b4}\cite{b5} primarily focus on utilizing RGB images captured by visible (VIS) cameras during the daytime, which are inadequate for accommodating 24-hour scenarios that involve infrared (IR) images captured by IR cameras. To address the substantial cross-modality gap and facilitate operation in all-day scenarios, visible-infrared person re-identification (VIReID) methods \cite{b6}\cite{b7} have been developed, enabling the matching of IR (RGB) images given an interest in a specific RGB (IR) pedestrian image.

\begin{figure}[t]
    \centering
    \subfigure[Features Learning with aligning cross-modalities knowledge]{
        \includegraphics[width=0.47\linewidth]{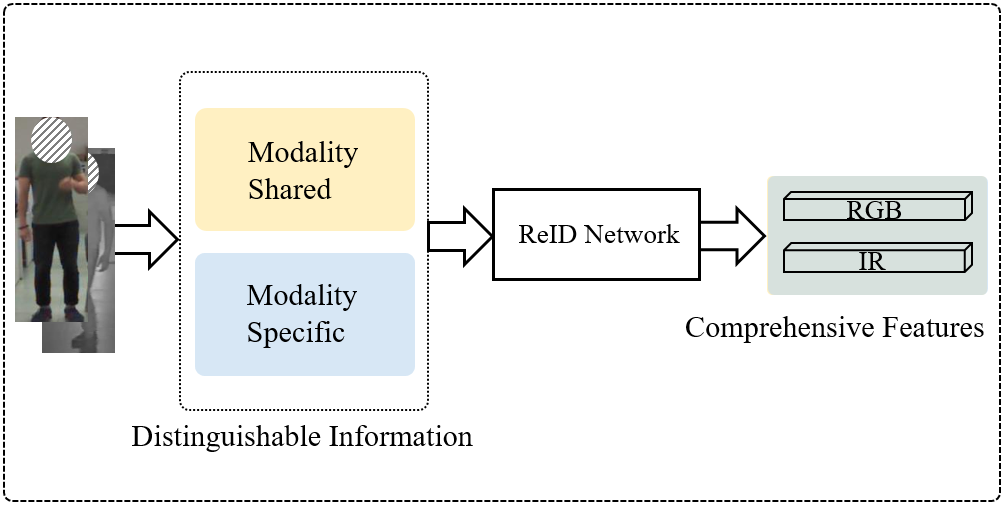}
    }
    \subfigure[Features Learning with cross-modalities knowledge compensation]{
        \includegraphics[width=0.47\linewidth]{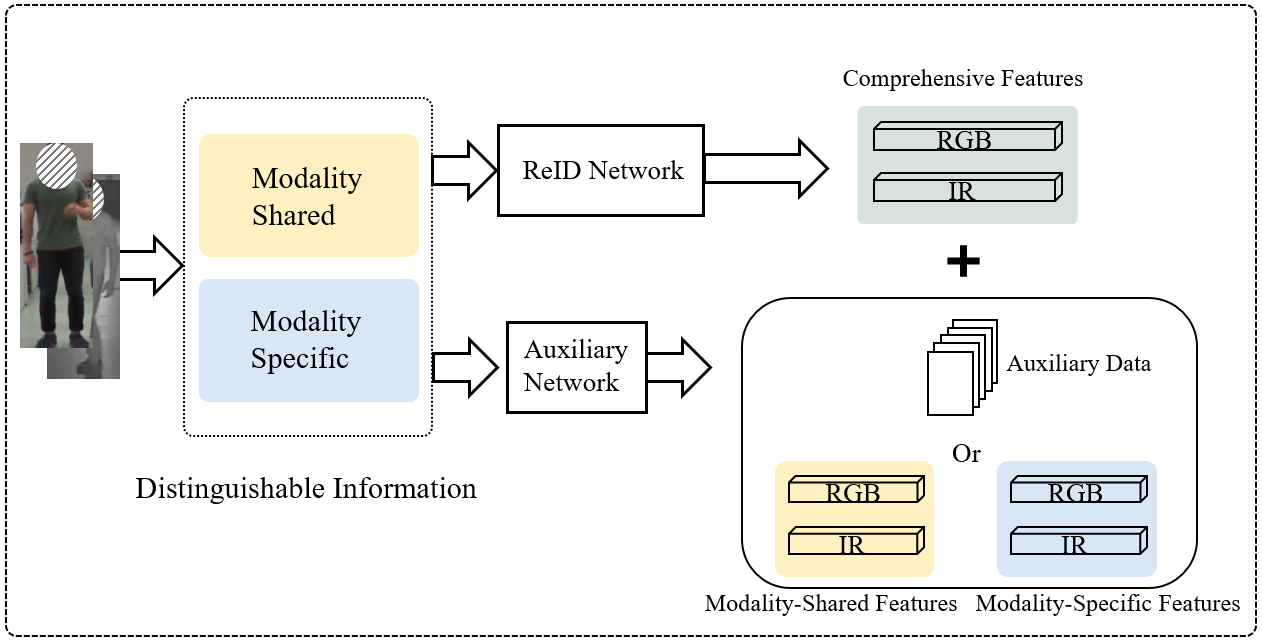}
    }
    \subfigure[Feature Learning with the proposed BDLF]{
        \includegraphics[width=0.7\linewidth]{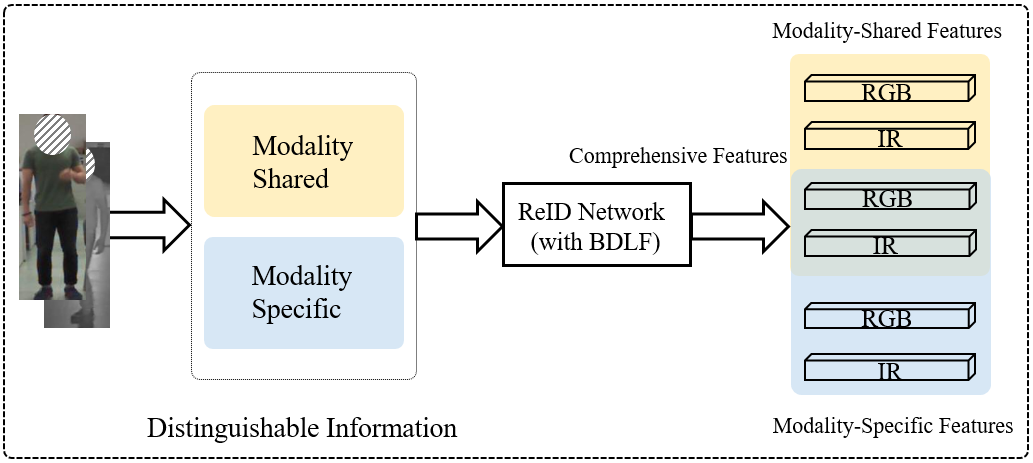}
    }
    \caption{Motivation of the proposed BDLF, which focuses on sufficiently mining the modality-shared and modality-specific knowledge simultaneously and are not applicable for additional auxiliary data.}
    \label{fig1}
\end{figure}

The existing research on VIReID can generally be categorized into two principal methods: extracting distinguishing modality-shared features from VIS and IR modalities\cite{b7}\cite{b8} and compensating for modality-specific or modality-shared features \cite{b9}. As shown in Figure \ref{fig1}(a), the former method aims to reduce cross-modality discrepancies by aligning comprehensive cross-modality features into a common semantic space. However, it neglects to leverage modality-specific and shared cues, which inevitably leads to performance bottlenecks. The latter approach, depicted in Figure \ref{fig1}(b) can be further divided into embedding-level and image-level methods. These methods generate compensatory knowledge in the embedding space and at the pixel level respectively, using auxiliary models(e.g., GANs\cite{b10}, segmentation networks, part alignment networks, etc.). However, these methods typically introduce losses and noise into the generated features or require additional data processing by other models, making them less effective and convenient. Consequently, advancing the development of VIReID to a more comprehensive level remains a significant challenge.

Inspired by the analyses presented above, it is essential to recognize that modality-shared information, such as the contour and movement characteristics of pedestrians, can be considered base features. In contrast, modality-specific information, including the color and texture details of the RGB modality and the thermal characteristics of the IR modality, can be regarded as detail features. Both types of them should be integrated and utilized  effectively together. Therefore, in this paper, we propose a novel Base-Detail Feature Learning Framework (BDLF), as shown in Figure \ref{fig1}(c). This framework is designed to extract modality-shared base features and modality-specific detail features from the original images with minimal additional computational costs, while jointly optimizing modality-shared, modality-specific, and comprehensive features.

 The proposed BDLF comprises a modality-specific detail feature extraction (DFE) module and a modality-shared base embedding generation (BEG) block, which ultimately combine the optimized features collected. Inspired by \cite{b17}, we designed the DFE module to mine the modality-specific detail information losslessly. Subsequently, the BEG block derives modality-shared base features. To fully capture both specific and shared information, we proposed a novel specific-shared knowledge distillation(SKD) loss. It encourages the detail (base) features to effectively incorporate modality-specific (modality-shared) knowledge by imposing a constraint on the correlation that the cross-modality detail and base features should exhibit. Specifically, it ensures that the correlations across RGB and IR modalities are indistinct and notable, respectively. Perspectives in \cite{b18} explain that the independent decomposition of features can maximize the mutual information of sub-features; therefore, we introduced an independence constraint in the semantic space between the derived detail and base features. This indicates that the base feature exclusively encompasses modality-shared knowledge, while the detail feature contains modality-specific information. In summary, the main contributions of our work are as follows:

\begin{itemize}
    \item A novel correlation optimization method is proposed that effectively generates both modality-shared and modality-specific features using a non-parametric approach, rather than relying on classifiers.
    \item We propose an end-to-end Base-Detail Feature Learning Framework (BDLF) for VIReID that integrates extracts of modality-shared base knowledge and modality-specific detail knowledge. 
    \item Extensive experiments have demonstrated that the proposed BDLF outperforms other state-of-the-art methods for the VIReID task on the SYSU-MM01, RegDB, and LLCM datasets.
\end{itemize}

\section{Related Work}
The main idea for solution VI-ReID task is decreasing the notable discrepence across VIS and IR modalities, thereby the existing methods consist of aligning the cross-modality features and utilizing the auxiliary data or features generated by other models. 

The alignment of feature representation methods seeks to convert cross-modality features into a unified semantic space through either metric learning techniques \cite{b2} \cite{b7} \cite{b16} or by enhancing networks with more effective feature extraction components \cite{b8} \cite{b34}. However, these approaches ultimately encounter performance bottlenecks due to the loss of modality-specific information.

The methods for utilizing auxiliary information produced by other models are proposed to enhance identifiable knowledge. GAN-based methods \cite{b9}d\cite{b36} generate compensatory  features at either the image level or the embedding level to simulate features from another modality. XIV \cite{b37} introduces the X-modality generated by a lightweight auxiliary network to decrease discrepancies between the two modalities. LUPI \cite{b31} establishes an intermediate domain between VIS and IR modalities. Furthermore, it generates images that belong to this intermediate domain to guide the network  in acquiring more discernible information. SGIEL \cite{b18} innovatively adopts the shape knowledge of identity generated by segmentation models to enrich supplementary information. TMD \cite{b29} generates style-aligned images to minimize differences at the image level, subsequently aligning cross-modality features to eliminate discrepancies in feature distribution and instance features. However, this remains a challenging field of research because these methods either inevitably introduce information distortion during the generation process or fail to completely capture modality-specific and modality-shared information.

\section{Methodology}


\begin{figure*}[t]
    \centering
    \begin{overpic}[width=1\linewidth]{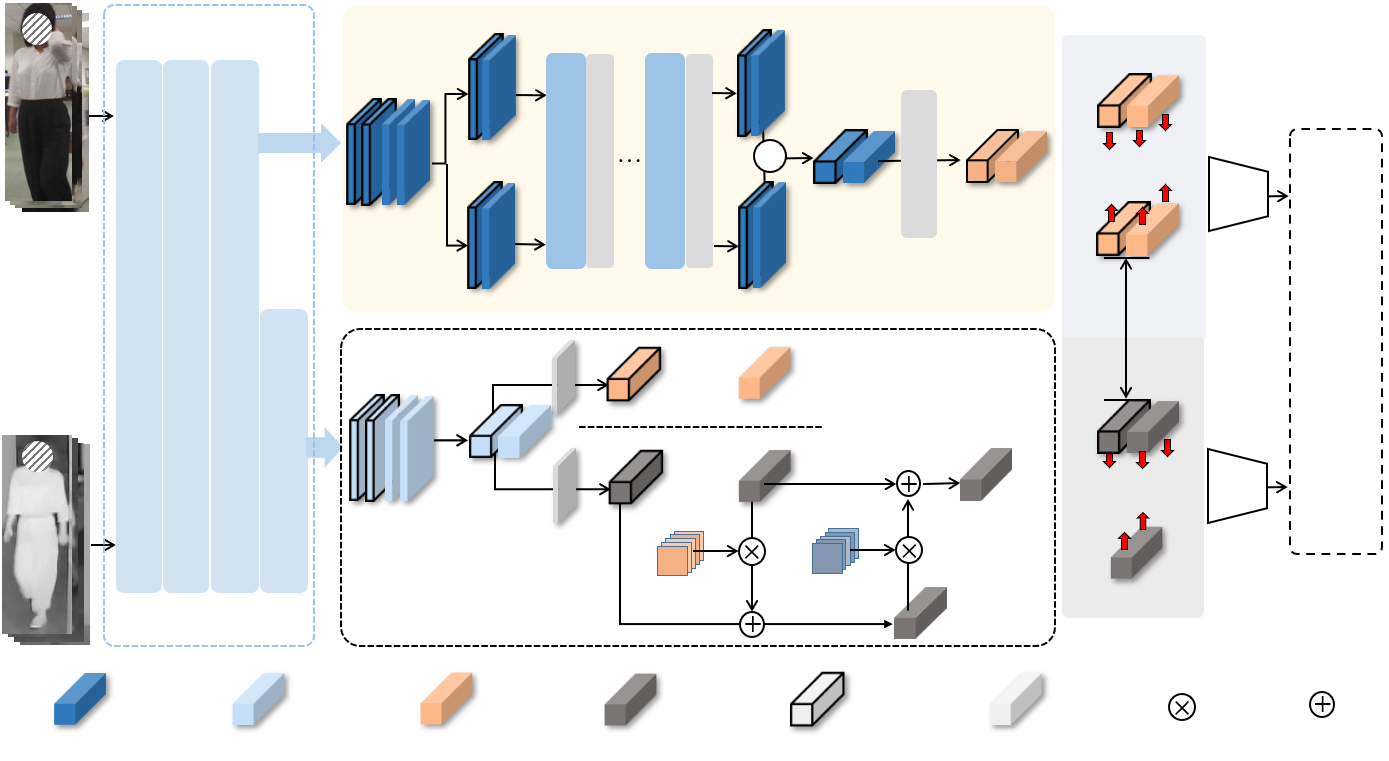}
        \put(9,22){\huge \rotatebox{90}{Conv Block 1}}
        \put(12.2,22){\huge \rotatebox{90}{Conv Block 2}}
        \put(15.5,22){\huge \rotatebox{90}{Conv Block 3}}
        \put(19.5,16.5){\large \rotatebox{90}{Conv Block 4}}
        \put(8,53){\small {Backbone Network}}
        
        \put(25,9){\small {Base Embedding Generation}}
        \put(27,16){\normalsize {$Z$}}
        \put(30,24){\small { $GAP$}}
        \put(34,18){\normalsize {$I-P$}}
        \put(36,28){\normalsize {$P$}}
        \put(31,15){\small {Projection matrix}}
        \put(49.5,28){\normalsize {$\bar Z^D$}}
        \put(49.5,20){\normalsize {$\bar Z^B$}}
        \put(42.5,22.5){\small {Independence Restrict}}
        \put(48,25){\small {$l_{orth}$}}
        \put(42.5,12){\small {Batch Attention}}
        \put(55,12){\small {Channel Attention}}
        \put(69,17){\normalsize {$Z^B_F$}}
        \put(67,9){\normalsize {$\bar Z^B_F$}}
        
        \put(25,53){\small {Detail Feature Extraction}}
        \put(27,38){\small {$Z^M$}}
        \put(22,34){\small {$Z^M_1(c+1:C)$}}
        \put(26,50.5){\small {$Z^M_1(1:c)$}}
        \put(56,34){\small {$Z^M_k(c+1:C)$}}
        \put(57,50.5){\small {$Z^M_k(1:c)$}}
        \put(54.3,43.2){\scriptsize  {$cat$}}
        \put(56,45){\small  {$Gap$}}
        \put(56.5,40){\small {$Z^M(1:C)$}}
        \put(71,40){\small {$Z^D$}}
        \put(40,39){\large \rotatebox{90}{INN Block}}
        \put(47,39){\large \rotatebox{90}{INN Block}}
        \put(42.5,35.8){\small \rotatebox{90}{Layer Normalization}}
        \put(49.5,35.8){\small \rotatebox{90}{Layer Normalization}}
        \put(65.5,37.5){\small \rotatebox{90}{Cross Attention}}
        
        \put(76,51){\small {Detail Subspace}}
        \put(76,11){\small {Base Subspace}}
        \put(76,37){\small {$Z^D$}}
        \put(76,47){\small {$\bar Z^D$}}
        \put(76,23){\small {$\bar Z^B$}}
        \put(77,13){\small {$Z^B_F$}}
        \put(80,19){\small {$l_{fbkl}$}}
        \put(80,42){\small {$l_{app}$}}
        \put(82,26){\scriptsize \rotatebox{90}{Correlation Restrict}}
        \put(79,30){\small \rotatebox{90}{$l_{corr}$}}

        \put(87,40.5){\scriptsize {$CLS_{D}$}}
        \put(87,19.5){\scriptsize {$CLS_{B}$}}

        \put(95,42){\small {$l_{id}$}}
        \put(95,37){\small {$l_{tri}$}}
        \put(95,32){\small {$l_{okl}$}}
        \put(94.5,27){\small {$l_{DFE}$}}
        \put(94.5,22){\small {$l_{BEG}$}}
        \put(95,17){\small {$l_{skd}$}}

        \put(1.5,1){\scriptsize {Middle Feature}}
        \put(12,1){\scriptsize {Comprehensive Feature}}
        \put(28,1){\scriptsize {Detail Feature}}
        \put(41,1){\scriptsize {Base Feature}}
        \put(55,1){\scriptsize {RGB Feature}}
        \put(70,1){\scriptsize {IR Feature}}
        \put(81,1){\scriptsize {Multiplication}}
        \put(92.5,1){\scriptsize {Addition}}
        
    \end{overpic}
    \caption{The pipeline of the proposed Base-Detail Feature Learning Framework (BDLF), which consists of a Detail Feature Extraction (DFE) module and a Base Embedding Generation (BEG) block, and jointly optimizes the extracted detail, base, and comprehensive features.}
    \label{fig2}
\end{figure*}
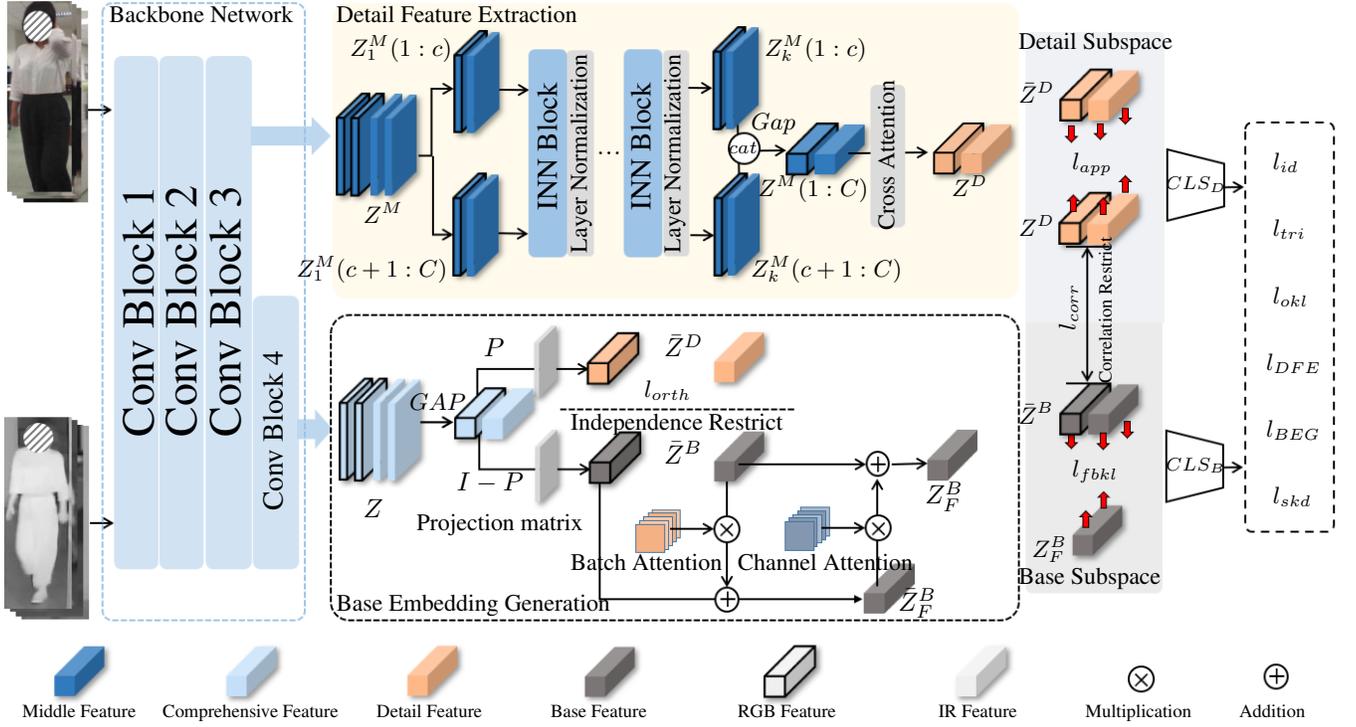

\subsection{Overall Framework}
The pipeline of our proposed method, referred to as BDLF, is illustrated in Figure \ref{fig2}. This method utilizes a single-stream ResNet-50 network\cite{b11} as its backbone. The intermediate features $Z^M\in\operatorname{R}^{B\times C\times H\times W}$, which pass through a portion of the backbone, are fed into the proposed detail feature extraction (DFE) module to yield detail features $Z^D$. Additionally, the base feature $Z^B$ is generated by inputting the output $Z\in\operatorname{R}^{\small B\times C}$ from the backbone into the proposed base embedding generation (BEG) block. A novel specific-shared knowledge distillation (SKD) loss is proposed to ensure that the generated detail(base) features contain as much modality-specific (modality-shared) knowledge as possible, thereby effectively leveraging modality-specific and shared information. Furthermore, we construct a modality-shared feature $Z^F$ using a cross-modality feature fusion method to optimally supplement the base features. During the inference phase, only the comprehensive feature $Z$ yielded by the backbone is used for performance evaluation. This is because the proposed DFE and BEG modules effectively enhance the comprehensive feature by incorporating additional detail and base information.

Given an identity image from either the visible or infrared modality, VIReID intends to identify the most similar sequence of that identity in another modality. Let the training set $\left\{X_V, X_I\right\}$ consist of $B$ identities, with each identity including $P$ samples. Therefore, $X_V = \left\{{{x_V}^{b,p}}, b = 1, ..., B;p = 1, ..., P\right\}$ symbolizes the set of visible images, while $X_I = \left\{{x_I}^{b,p}, b = 1, ..., B;p = 1, ..., P\right\}$ denotes the set of infrared images. As illustrated in Figure \ref{fig2}, the VIS and IR images are processed through the backbone network, i.e,
\begin{align}
     Z_{V/I}^M = &E^{fore}(X_{V/I})\nonumber \\ 
     Z_{V/I} = &E^{rear}(Z_{V/I}^M)\nonumber \\ 
     Z = c&at(Z_V,Z_I)\label{f1}
\end{align}

where $E^{fore}(\cdot)$ and $E^{rear}(\cdot)$ are the former and latter parts of the backbone network, the embeddings ${Z_{V/I}}^M\in\operatorname{R}^{\small \frac{B}{2}\times C\times H\times W}$ and $Z_{V/I}\in\operatorname{R}^{\small \frac{B}{2}\times C}$ denote the intermediate and complete outputs from the backbone for the VIS and IR modalities, $cat(\cdot)$ refers to the concatenation operation along the batch dimension.

\subsection{Specific-shared Knowledge Distillation}
\begin{figure}[t]
    \centering
    \begin{overpic}[width=0.99\linewidth]{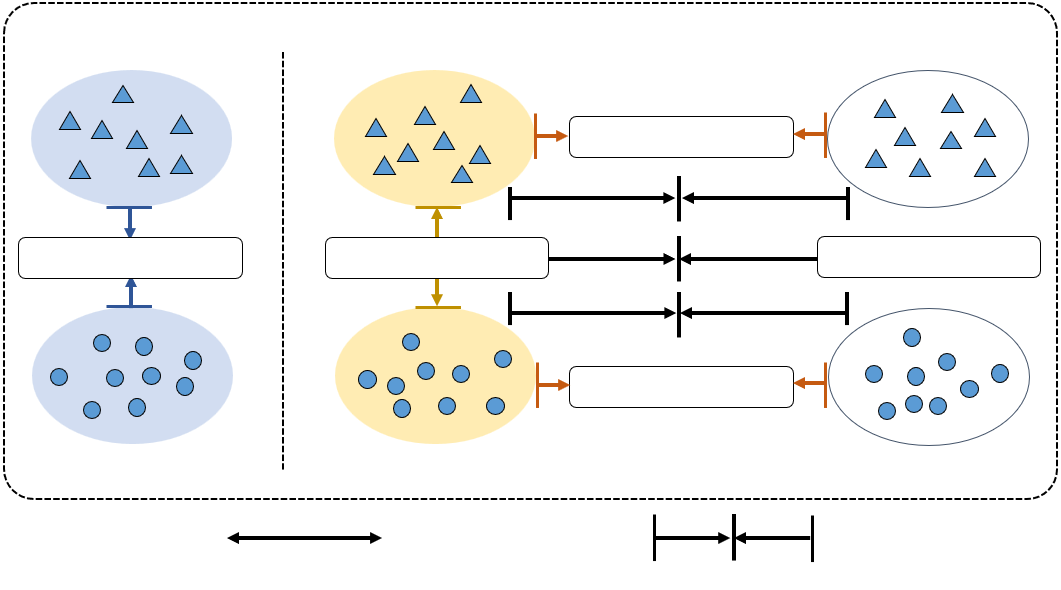}
    
        \put(2,10){\scriptsize {IR Base Embeddings}}
        \put(1,52){\scriptsize {VIS Base Embeddings}}
        \put(30,10){\scriptsize {IR Detail Embeddings}}
        \put(29,52){\scriptsize {VIS Detail Embeddings}}
        \put(59,10){\scriptsize {IR Detail Embeddings From BEG}}
        \put(58,52){\scriptsize {VIS Detail Embeddings From BEG}}

        \put(2,30.5){\scriptsize {$Corr(Z^B_V,Z^B_I)$}}
        \put(30.5,30.5){\scriptsize {$Corr(Z^D_V,Z^D_I)$}}
        \put(77,30.5){\scriptsize {$Corr(\bar Z^D_V,\bar Z^D_I)$}}
        \put(53.5,42){\scriptsize {$Corr(Z^D_V,\bar Z^D_V)$}}
        \put(53.5,18.5){\scriptsize {$Corr(Z^D_V,\bar Z^D_V)$}}

        \put(23.5,1){\scriptsize {Pull away}}
        \put(64,1){\scriptsize {Pull close}} 
        
    \end{overpic}
    \caption{Illustration of correlation instruct to learn modality specific and shared information.}
    \label{fig3}
\end{figure}
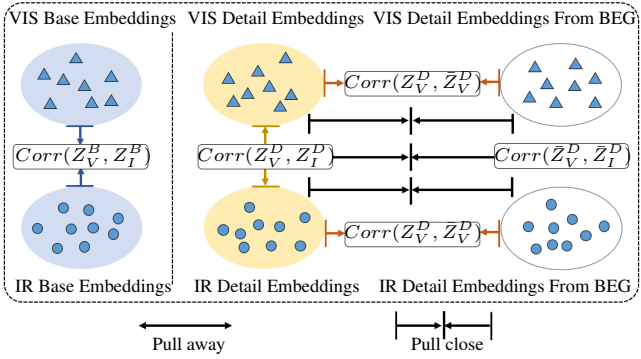

 We observe that the similarity of base information, such as contours and movements, between the VIS and IR modalities is noticeable. In contrast, the similarity of detail information including color, texture, and thermal details between the two modalities is suppressed. Inspired by \cite{b17}, as shown in Figure \ref{fig3}, the base and detail features can be generated by increasing and reducing the correlation between the two modalities respectively. Based on this, we propose a novel specific-shared knowledge distillation (SKD) loss, which is numerically smoother and easier to optimize, formulated as follows:
\begin{align}
    l_{skd} =  \dfrac{log[Corr(Z_V^B,Z_I^B)]}{\sqrt[3]{log[Corr(Z_V^D,Z_I^D)]}+\gamma}\label{skd}
\end{align}
in which $Z_{V/I}^B$ denotes the base features generated by the proposed BEG block, and  $Z_{V/I}^D$ denotes the detail features extracted from the proposed DFE module.
$Corr(\cdot)$ is the Pearson correlation coefficient operation, while $\gamma$ represents a constant that ensures the denominator remains non-zero. According to optimize the SKD loss, the correlation between the VIS and IR modalities of both base and detail features(i.e, $Corr(Z_V^B,Z_I^B)$ and $Corr(Z_V^D,Z_I^D)$ in formula (\ref{skd})) is simultaneously increased and decreased. This approach allows the proposed DFE module to extract embeddings rich in detailed knowledge. Consequently, the proposed BEG block is capable of generating base embeddings that contain a greater amount of modality-shared knowledge.


\subsection{Detail Feature Extraction}
The proposed DFE module aims to acquire detail features that imply modality-specific information from the intermediate embedding $Z_{V/I}^M$ by utilizing a series of invertible neural network (INN) blocks\cite{b17}\cite{b12}\cite{b13}, which can effectively preserves detailed characteristics and mitigates information loss during feature extraction by making its input and output embeddings are mutually generated. Taking the VIS case as an example, we obtain the input for the DFE module, ${Z_V(1:c)}^M$ and ${Z_V(c+1:C)}^M\in\operatorname{R}^{\frac{B}{2}\times \frac{C}{2}\times H\times W}$ by splitting ${Z_V}^M$ in half along the channel dimension. The transformations in each block can be denoted as follows:
\begin{align}
     &Z_{V,k+1}^M(c+1:C) = Z_{V,k}^M(c+1:C) + F_1[Z_{V,k}^M(1:c)]\nonumber \\ 
     &Z_{V,k+1}^M(1:c) = F_2[Z_{V,k+1}^M(c+1:C)]\nonumber \\  
     &\qquad\qquad\quad+ Z_{V,k}^M(1:c)\bullet exp\{F_3[Z_{V,k+1}^M(c+1:C)]\}\nonumber \\
     &Z_{V,k+1}^M = LN\{cat[Z_{V,k+1}^M(1:c),Z_{V,k+1}^M(c+1:C)]\}
\end{align}
Here, $Z_{V,k}^M$ is the input of the $k$th ($k\in1,...,K$) block, $F_i(\cdot)(i\in1,2,3)$ denotes the convolution blocks. The symbol $\bullet$ indicates element-wise multiplication of matrices, $LN(\cdot)$ represents layer normalization in the Lite Transformer\cite{b14} and $cat(\cdot)$ is the channel concatenation operation. The IR situation can be easily derived by substituting $I$ for the subscript $V$ in the aforementioned formulas.

At the final stage of DFE, we consider the detail features of both modalities integrally, as concatenating the extracted detail embeddings from the two modalities can help reduce computational complexity. Therefore, we feed the extracted detail embeddings into a cross-attention-based transformer to facilitate cross-modality reasoning and information exchange. This process enables the detail feature integration of knowledge from various modalities and allows for a more effective focus on distinguishable information, thereby enhancing the robustness and efficacy of semantic representation. Inspired by\cite{b15}, the transformer can be denoted as follows:
\begin{align}
    &\qquad\qquad\qquad Z_{V/I}^P = GAP(Z_{V/I,K}^M)\nonumber \\
    Z_V^D &= LN\{softmax[(Z_V^P W_q)(Z_I^P W_k)^T](Z_I^P W_v)+Z_V^P\}\nonumber \\ 
    Z_I^D &= LN\{softmax[(Z_I^P W_q)(Z_V^P W_k)^T](Z_V^P W_v)+Z_I^P\}
\end{align}
where $GAP(\cdot)$ is the global average pooling operation, $Z_{V/I}^P\in\operatorname{R}^{\frac{B}{2}\times \frac{C}{2}}$ denotes the embeddings after pooling. $W_q$, $W_k$ , and $W_v$ are the learnable parameters for DFE, $LN(\cdot)$ refers to layer normalization, and $softmax(\cdot)$ indicates the calculation of the softmax by row. Ultimately, the detail feature $Z^D\in\operatorname{R}^{B\times \frac{C}{2}}$ produced by the proposed DFE module is obtained by concatenating the VIS and IR detail embeddings along the batch dimension:
\begin{align}
    Z^D = cat(Z_V^D,Z_I^D)
\end{align}

With the proposed SKD loss formulated in formula (\ref{skd}), the extracted detail feature $Z^D$ can significantly enrich modality-specific detail knowledge. Thus a private classifier $CLS_D$ that is specially designed for the detail feature $Z^D$ is constructed, alongside a communal classifier $CLS_B$ that processes the base embeddings and the comprehensive feature $Z$ obtained from formula (\ref{f1}), as illustrated in Figure \ref{fig2}. Furthermore, the commonly used id loss\cite{b16} driven by cross-entropy ($ce(p,q) = -	\sum_{i=1}^{n} q_i log(p_i)$) was applied to strengthen the distinguishable information of detail feature $Z^D$, i.e,
\begin{align}
    l_{id}^D = \operatorname{E}_{(z^D \sim Z^D)} ce(CLS_D(z^D),Y) \label{idd}
\end{align}

Since there are differences in the distribution of classification results between the detail feature $Z^D$ and the comprehensive feature $Z$, this misalignment may impede our goal of enhancing the representation ability of $Z$ leveraging detailed knowledge. Therefore, we constrain the probability distribution predicted from $Z^D$ to align with the distribution from $Z$, ensuring that their semantic representations are consistent. This process can be expressed as follows:
\begin{align}
    l_{odkl} = \operatorname{E}_{(z,z^D \sim Z,Z^D)}ce(CLS_D(z^D),CLS_B(z)) \label{od}
\end{align}

The total loss of the proposed DFE module can be obtained by combining formulas (\ref{idd}) and  (\ref{od}):
\begin{align}
    l_{DFE} = l_{id}^D + l_{odkl}
\end{align}

\subsection{Base Embedding Generation}
The proposed BEG block is designed to produce the base embeddings from $Z$ utilizing the acquired detail feature $Z^D$. Take notice that there are significant semantic differences between modality-specific detail information such as color and texture and modality-shared base information, which includes movements, contours, and so on. For this reason, inspired by \cite{b18}, we have developed a method to ensure that the detail(base) features can only contain modality-specific(modality-shared) distinguishable knowledge, thereby maximizing the collection of both modality-specific and modality-shared information. Furthermore, the proposed DFE and BEG blocks can learn these two categories of knowledge simultaneously without interfering with each other. Based on this premise, we consider the detail and base embeddings to be independent of each other, i.e, $Z^D \perp Z^B$. According to the approach of making $\bar Z^D$ comprehensively converge to $Z^D$ and impose the independence restriction between the detail and base embedding, the proposed BEG block can then generate modality-shared base embedding by excluding detailed knowledge from $Z$ in the semantic space,i.e,
\begin{equation}
    \begin{cases}
        Z\times P = \bar Z^D&,\bar Z^D \rightarrow Z^D \\
        Z\times (I-P) = Z^B& 
    \end{cases}
\end{equation}
in which $Z$ is the output of backbone network, $I$ is the identity matrix, $\rightarrow$ denotes approximating, $\bar Z^D$ and $Z^B\in\operatorname{R}^{B\times C}$ are the gained detail and base embeddings by using a projection matrix $P\in\operatorname{R}^{C\times C}$ to decompose $Z$ into mutually orthogonal subspaces. By the properties of orthogonal projection matrix, $P$ should be a conjugate symmetric idempotent matrix and must satisfy the following constraints in the real number case:
\begin{align}
    P^2 = P,P^T = P \label{orth}
\end{align}

The process of approaching can be divided into three components: approximating in the feature space, semantic representation, and the correlation between $Z^D$ and $\bar Z^D$. In the case of approximating on feature space, we first calculate the distances between all embeddings in a mini-batch for $\bar Z^D$ and $Z^D$ respectively, and obtain the difference map $M$ by:
\begin{align}
    M = ||softmax&[\bar Z^D(\bar Z^D)^T - Z^D(Z^D)^T]||^2
\end{align}
Then we enforce the distance distribution of $\bar Z^D$ to converge to that of $Z^D$ by optimizing the following loss:
\begin{align}
    l_{fkl} &= \operatorname{E}_{(a_{i,j}\sim M)}a_{i,j}
\end{align}
Furthermore, we aligned the semantic representation between $Z^D$ and $\bar Z^D$ by adjusting the predicted probability distribution of $\bar Z^D$ closer to that of $Z^D$. By drawing an analogy with formula (\ref{od}), we have:
\begin{align}
    l_{dkl} = \operatorname{E}_{({\bar z}^D,z^D \sim {\bar Z}^D,Z^D)}ce(CLS_D({\bar z}^D), CLS_D(z^D))
\end{align}

Considering that the detail feature $\bar Z^D$ generated by the BEG block should exhibit the same correlation properties as $Z^D$. As illustrated on the right side of the dashed line in Figure \ref{fig3}, we achieved consistency in correlation between $\bar Z^D$ and $Z^D$ by pulling close their cross-modalities correlations denoted as $Corr(Z^{p}_{V},Z^{p}_{I}),p \in \{D,/\bar D \}$ and by reducing the discrepancy in correlation within the same modality, represented as $Corr(Z^{D}_{m},Z^{\bar D}_{m}),m \in \{V,I\}$. This is accomplished by optimizing the follows loss:
\begin{align}
    l_{dcorr} = \frac{(Corr({\bar Z}_V^D,{\bar Z}_I^D)-Corr(Z_V^D,Z_I^D))^2}{Corr({\bar Z}_V^D,Z_V^D)^2 + Corr({\bar Z}_I^D,Z_I^D)^2 + \gamma}
\end{align}
Thereby, the total approaching function for $\bar Z$ is:
\begin{align}
    l_{app} = l_{fkl} + l_{dkl} + l_{dcorr}
\end{align}

After the description provided above, we generated the base feature $Z^B$ by eliminating the detail feature $Z^D$ from $Z$. Given that the base information across modalities, such as contours and movements, should exhibit significant similarities, we constructed a cross-modality feature fusion method that integrates the base feature $Z_V^B$ and $Z_I^B$ to generate an auxiliary feature $Z_F^B$. Inspired by \cite{b15} \cite{b19}, the fusion method can be formulated as follows:
\begin{align}
    \bar Z_F^B &= \frac{1}{C}[(Z_V^B P_q)^T(Z_I^B P_k)](Z_I^B P_v)+Z_V^B\nonumber \\ 
    Z_F^B &= \frac{2}{B}[(Z_I^B Q_q)(\bar Z_F^B Q_k)^T](\bar Z_F^B Q_v)+Z_I^B
\end{align}
Here, $Z_{V/I}^B \in \operatorname{R}^{\frac{B}{2} \times C}$ represents the cross-modality base embedding, $P$,$Q$ are the learnable parameters. The fused $Z_F^B$ aggregates the base knowledge from VIS and IR modalities, employing attention mechanisms across both channel and batch dimensions. We then enhance the similarity between $Z_V^B$ and $Z_I^B$ by aligning them with $Z_F^B$:
\begin{align}
    l_{fbkl} = \operatorname{E}_{(z_F^B,z_{V/I}^B \sim Z_F^B,Z_{V/I}^B)}ce(CLS_B(z_{V/I}^B),CLS_B(z_F^B))
\end{align}
This approach ensures that $Z^B$ contains only the knowledge shared between the modalities. In addition, we also utilize cross-modality semantic alignment for $Z_{V/I}^B$ to strengthen the collection of modality-shared knowledge:
\begin{align}
    l_{bkl} = \operatorname{E}_{(z_{V/I}^B \sim Z_{V/I}^B)}ce(CLS_B(z_V^B),CLS_B(z_I^B)) \label{bkl}
\end{align}
The id loss for both was also employed to enhance the distinguishable information of $Z^B$ and $Z_F^B$, and the loss for cross-modality feature fusion method is:
\begin{align}
    l_{cmf} = l_{id}^F + l_{fbkl}
\end{align}
Consequently, the total loss for the BEG block can be summarized as follows:
\begin{align}
    l_{BEG} = l_{id}^B + l_{app} + l_{bkl} + l_{cmf} + l_{orth}
\end{align}
where $l_{orth}$ represents the constraint in formula (\ref{orth}) for parameter $P$ to achieve the decomposition of orthogonal subspaces.
\subsection{Optimization}
In the preceding section, the proposed DFE module extracted detailed knowledge from the intermediate feature $Z^M$ and subsequently produced the detail feature $Z^D$, the proposed BEG block produced the base feature by eliminating the detailed knowledge from the comprehensive feature $Z$, the proposed SKD loss ensures that both the detail and base features effectively capture modality-specific and shared information. We also incorporated the commonly used id and triplet loss \cite{b20} $l_{tri}$ for $Z$ into our method. Similar to (\ref{bkl}), we enforce cross-modality consistency for $Z$ by:
\begin{align}
    l_{okl} = \operatorname{E}_{(z_{V/I} \sim Z_{V/I})}ce(CLS_B(z_V),CLS_B(z_I)) \label{okl}
\end{align}

Eventually, the total loss of BDLF is defined as:
\begin{align}
    l_{totel} = l_{id} + l_{tri} + l_{okl} + l_{DFE} + l_{BEG} + l_{skd}
\end{align}

\section{Experiments}
In this section, we validate the effectiveness of our BDLF by conducting experiments on the widely recognized SYSU-MM01, RegDB and LLCM benchmarks.

\begin{table*}[t]
    \centering
    \tabcolsep=0.08cm
    \begin{tabular}{ l l r r r r r r r r r r r r } 
        \toprule
        \multirow{3}*{Methods}&\multirow{3}*{Venue}&\multicolumn{4}{c}{SYSU-MM01}&\multicolumn{4}{c}{RegDB}&\multicolumn{4}{c}{LLCM}\\
        \cmidrule {3-14}
        ~&~&\multicolumn{2}{c}{All-Search}&\multicolumn{2}{c}{Indoor-Search}&\multicolumn{2}{c}{VIS to IR}&\multicolumn{2}{c}{IR to VIS}&\multicolumn{2}{c}{VIS to IR}&\multicolumn{2}{c}{IR to VIS}\\
        \cmidrule {3-14}
        ~&~&R-1&mAP&R-1&mAP&R-1&mAP&R-1&mAP&R-1&mAP&R-1&mAP\\
        \midrule 
        CAJ\cite{b30}&ICCV'21&69.9&66.9&76.3&80.4&85.0&79.1&84.8&77.8&56.5&59.8&48.8&56.6\\
        MMN\cite{b27}&ACMMM'21&70.6&66.9&76.2&79.6&91.6&84.1&87.5&80.5&59.9&62.7&52.5&58.9\\  
        FMCNet\cite{b9}&CVPR'22&66.3&62.5&68.2&74.1&89.1&84.4&88.4&83.9&-&-&-&-\\
        LUPI\cite{b31}&ECCV'22&71.1&67.6&82.4&82.7&88.0&82.7&86.8&81.3&-&-&-&-\\
        MSCLNet\cite{b28}&ECCV'22&\underline{77.0}&71.6&78.5&81.2&84.2&81.0&83.7&78.3&-&-&-&-\\
        DEEN\cite{b8}&CVPR'23&74.7&71.8&80.3&83.3&91.1&85.1&89.5&83.4&62.5&65.8&54.9&62.9\\
        SGIEL\cite{b18}&CVPR'23&75.2&70.1&78.4&81.2&92.2&86.6&91.1&85.2&-&-&-&-\\
        TMD\cite{b29}&TMM'23&73.9&67.8&81.2&78.9&93.0&84.3&87.4&81.3&-&-&-&-\\
        AGCC\cite{b33}&PR'24&75.9&73.0&79.3&84.6&92.6&86.2&91.4&84.9&-&-&-&-\\
        ReViT\cite{b34}&PR'24&68.1&65.1&72.4&77.6&91.7&86.0&93.0&86.1&-&-&-&-\\
        STAR\cite{b32}&TMM'24&76.1&72.7&83.5&\underline{85.8}&94.1&88.8&93.3&88.2&-&-&-&-\\
        \midrule
        BDLF(ours)&-&76.8&\underline{74.6}&\underline{84.2}&\underline{85.8}&\underline{94.4}&\underline{90.1}&\underline{94.5}&\underline{89.6}&\underline{67.0}&\underline{68.9}&\underline{58.1}&\underline{64.5}\\
        \bottomrule         
    \end{tabular}
    \caption{Comparisons between the proposed BDLF and several state-of-the-art methods on the SYSU-MM01, RegDB, and LLCM datasets.}
    \label{tab_1}
\end{table*}


\begin{table}[t]
    \centering
    \tabcolsep=0.28cm
    \begin{tabular}{ c c c c c r r } 
        \toprule
        \multicolumn{5}{c}{Settings}&\multicolumn{2}{c}{SYSU-MM01}\\
        \midrule 
        DFE&BEG&$l_{app}$&$l_{orth}$&$l_{skd}$&R-1&mAP\\
        \midrule 
        & & & & &-&- \\
        $\checkmark$& & & & &-&- \\  
        &$\checkmark$& & & &72.7&68.1 \\
        $\checkmark$&$\checkmark$& & & &73.7&69.6 \\
        $\checkmark$&$\checkmark$&$\checkmark$& & &75.3&72.1 \\
        $\checkmark$&$\checkmark$& & &$\checkmark$&73.7&69.0 \\
         &$\checkmark$&$\checkmark$&$\checkmark$&$\checkmark$&74.0&70.9 \\
        $\checkmark$&$\checkmark$&$\checkmark$& &$\checkmark$&75.5&72.3 \\
        $\checkmark$&$\checkmark$&$\checkmark$&$\checkmark$&$\checkmark$&76.8&74.6 \\
        \bottomrule         
    \end{tabular}
    \caption{Effectiveness of each component for the proposed BDLF.}
    \label{tab_2}
\end{table}

\subsection{Datasets and Evaluation Protocol}
SYSU-MM01 dataset \cite{b21} comprises 287,628 VIS and 15,792 IR images from 491 identities captured by 4 RGB and 2 IR cameras.  It features both All-Search and Indoor-Search modes for evaluation. RegDB \cite{b22} contains 412 identities, each represented by 10 VIS and 10 IR images captured from a pair of cameras. We adhere to the evaluation protocol outlined in \cite{b23} to randomly split the identities into training and testing sets of equal size. LLCM \cite{b8}is a challenging large-scale low-light dataset for VI-ReID task, which contains 713 identities with 25,626 VIS and 21,141 IR images, all captured by 9 cameras in both RGB and IR modalities 

The Cumulative Matching Characteristic curve (CMC) and mean Average Precision(mAP) are adopted as standard evaluation metrics in our experiments to comprehensively assess the performance of our framework.

\subsection{Implementation Details}
The entire framework is implemented using PyTorch and runs on a single NVIDIA RTX3090 GPU with 24GB VRAM. We employed a pre-trained ResNet-50\cite{b24} as the backbone network and incorporated INN blocks with affine coupling layers\cite{b12}\cite{b13} to construct the DFE module, setting the number of INN blocks to 6. All images are resized to $3\times384\times144$, and we adopted the Random Channel Exchangeable Augmentation and Channel-Level Random Erasing techniques proposed in \cite{b25} during the training phase. The SGD optimizer was used, with the initial learning rate set to $1\times10^{-2}$, which was warmed up to $1\times10^{-1}$ during the first 10 epochs, then we decayed the learning rate to $1\times10^{-2}$ and $1\times10^{-3}$ at epochs 20 and 95 for SYSU-MM01, and at epochs 70 and 130 for RegDB and LLCM, respectively. The learning rate was further decayed to $1\times10^{-4}$ at 180 epoch, with a total of 220 epochs. For each mini-batch, we randomly sampled 8 identities, each consisting of 4 VIS and 4 IR images for training. Additionally, the exponential moving average (EMA) model \cite{b26} also employed in our method.

\subsection{Comparison with State-of-the-art Methods}
We demonstrate the superiority of our BDLF by comparing performance with several existing state-of-the-art methods on the SYSU-MM01, RegDB, and LLCM datasets. The performance of these methods is presented in Table \ref{tab_1}, with optimal performances annotated by underlining.

\paragraph{Comparison on SYSU-MM01 and RegDB.}Table \ref{tab_1} presents the results of our BDLF alongside selected outstanding methods, confirming the superiority of our BDLF, which almost outperforms all other state-of-the-art methods. In the All-Search mode of SYSU-MM01, our method achieved a rank-1 accuracy of $76.8\%$ and a mAP of $74.6\%$, in the Indoor-Search mode, BDLF achieved a rank-1 accuracy of $84.2\%$ and a mAP of $85.8\%$. On the RegDB dataset, our method achieved a rank-1 accuracy of $94.4\%$ and a mAP of $90.1\%$ for the VIS to IR search, and attained a rank-1 accuracy of $94.5\%$ and a mAP of $89.6\%$ for the IR to VIS search. These results validate the effectiveness of BDLF that independently learns the detail and base information and sufficiently utilizes cross-modalities knowledge.

\paragraph{Comparison on LLCM.} According to Table \ref{tab_1}, our method outperformed other approaches. Specifically, BDLF achieved a rank-1 accuracy of $67.0\%$ and a mAP of $68.9\%$ in VIS to IR search, as well as a rank-1 accuracy of $58.1\%$ and a mAP of $64.5\%$ in IR to VIS search. It is evident that our BDLF is well-equipped to handle challenging scenarios.

\begin{figure}[t]
    \centering
    \subfigure{
        \includegraphics[width=1\linewidth]{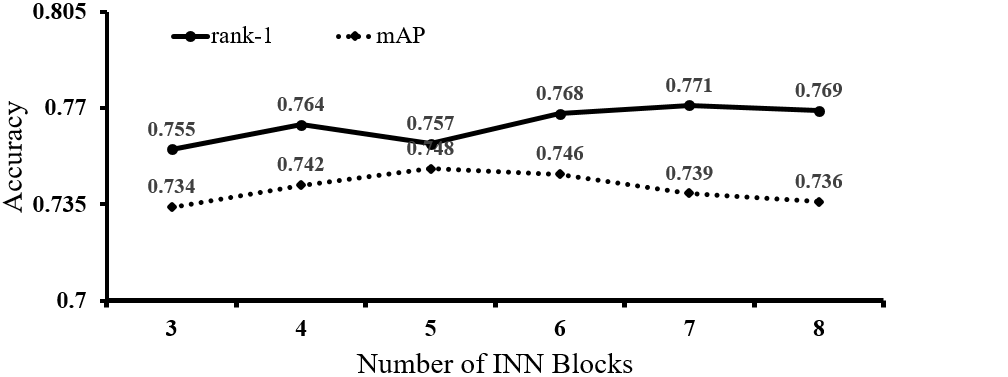}
    }
    \caption{Effectiveness of how many INN blocks are more favorable for the proposed DFE.}
    \label{fig4}
\end{figure}

\begin{table}[t]
    \centering
    \tabcolsep=0.65cm
    \begin{tabular}{ c r r } 
        \toprule
        \multirow{2}*{Location of DFE}&\multicolumn{2}{c}{SYSU-MM01}\\
        \cmidrule{2-3}
        \multirow{2}*{~}&R-1&mAP\\
        \midrule
        After stage-1&59.9&55.3\\
        After stage-2&71.1&67.1\\
        After stage-3&76.8&74.6\\
        After stage-4&73.4&71.1\\
        \bottomrule
    \end{tabular}
    \caption{Effectiveness of which stage of ResNet-50 to combine the proposed DFE.}
    \label{tab_3}
\end{table}

\begin{figure}[t]
    \centering
    \subfigure{
        \includegraphics[width=1\linewidth]{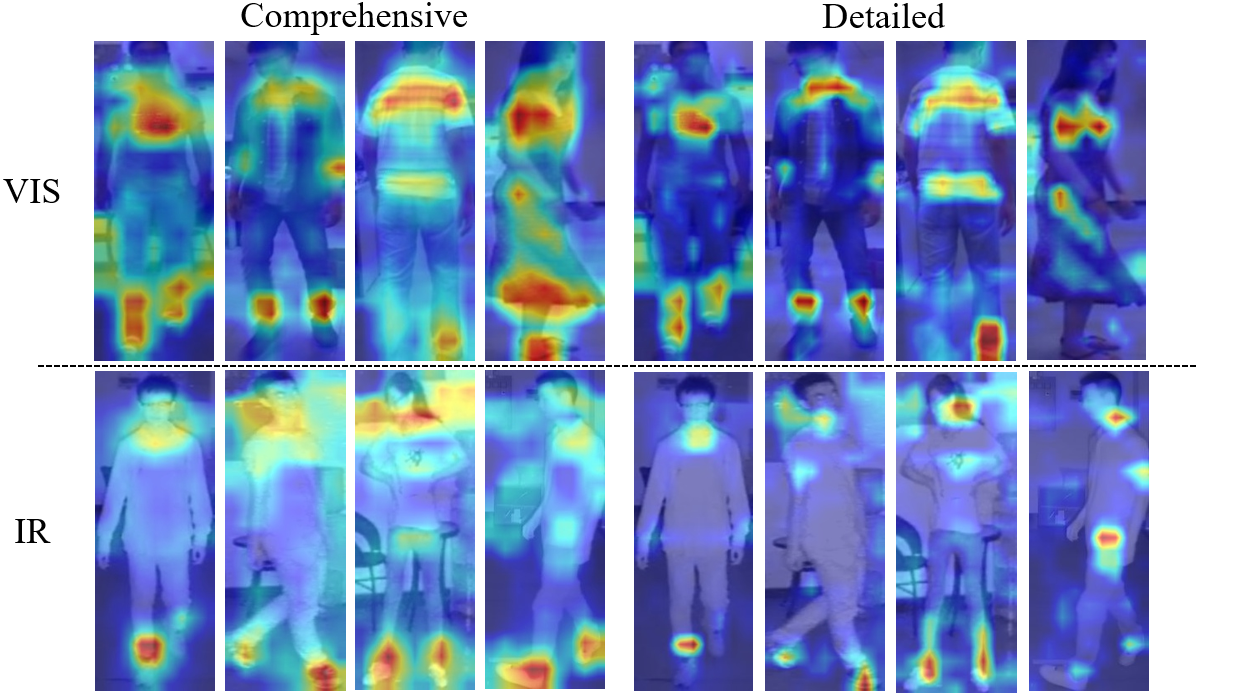}
    }
    \caption{Visualization of the comprehensive and detailed features.}
    \label{fig5}
\end{figure}

\subsection{Ablation Studies}
\paragraph{Effectiveness of each component.} In this section, we designed an ablation experiment to validate the effectiveness of certain components of BDLF. Specifically, we removed the DFE,  $l_{app}$, $l_{orth}$ and $l_{skd}$ modules from BDLF, while retaining the backbone with the BEG block as the baseline. All experiments adopted the same training settings, and we evaluated their performance in the All-Search mode of SYSU-MM01. The results are presented in Table \ref{tab_2}, Notably, the removal of the DFE module resulted in poor precision, demonstrating the effective detail extraction capability of DFE. The experiments also indicated that the $l_{app}$ loss enhances the model's distinguishing performance by effectively aiding in the generation of base embeddings, eliminating detailed knowledge from the comprehensive feature. Although the DFE module significantly promotes the mining of detail information, its performance remains suboptimal, as the model cannot extract all modality-specific and shared information without interference each other due to the absence of correlation constraint $l_{corr}$ and independent constraint $l_{orth}$ .

\paragraph{Effectiveness of how many INN blocks are more favorable for DFE.} The proposed DFE module consists of a series of INN blocks with an LN layer to extract detail information non-destructively. We conducted experiments to determine the optimal number of blocks for our framework. As shown in Figure \ref{fig4}, we modified the number of INN blocks and evaluated performance in the All-Search mode of SYSU-MM01. The results indicate accuracy gradually improves as the number of INN blocks increases, reaching a plateau when the count is 6. This confirm that a balance exists between accuracy and computational complexity when the number of INN blocks is set to 6.

\paragraph{Effectiveness of which stage of ResNet-50 to combine DFE module.} In this section, we implement experiments to assess which stage of ResNet-50 is most suitable for serving as the input to the proposed DFE module. All experiments maintain consistent settings, except for the locations of the DFE module within ResNet-50. The results are presented in Table \ref{tab_3}, we observed that connecting the DFE module to stage-3 of ResNet-50 yielded the best accuracy in the All-Search mode of SYSU-MM01. This can be attributed to the fact that modality-shared information is more prominent in the high-level features produced by stages-4, which impedes the extraction of modality-specific detail information. Furthermore, the low-level features  generated by stages 1 and 2 are inadequate for effectively expressing the semantics necessary to distinguish between different identities. These findings elucidate why the best accuracy is achieved when the DFE module is connected to stage-3 of ResNet-50.

\subsection{Visualization}
To investigate the detail information extraction capabilities of the proposed DFE, we visualize the comprehensive and detailed features of several identities produced by BDLF. As illustrated in Figure \ref{fig5}, a comparison of the images of comprehensive and detailed features reveals that the attention regions of the comprehensive features is broader and more dispersed than that of the detailed features. This observation indicates that the DFE module has the capacity to focus on subtly distinguishable characteristics.
\section{CONCLUSION}
In this paper, we propose a novel base-detail feature learning framework(BDLF) that learns detail and base features from a correlation and mutual information maximization for the VI-ReID task. The proposed BDLF consists of a DFE module and a BEG block. The DFE module non-destructively extracts detail information, while the BEG block generates base features by eliminating detail information from the output of the backbone network. By imposing constraints of independence and correlation on the detail and base embeddings, the proposed BDLF can capture detail and base features that retain as much modality-specific and shared information as possible, thereby effectively leveraging the differentiated minutiae. Extensive experiments on the SYSU-MM01, RegDB, and LLCM datasets have demonstrated the superiority of BDLF.

\section*{Acknowledgments}
This work was partially supported by the Shenzhen Science and Technology Program (KJZD20230923114600002), and the Guangdong Major Project of Basic and Applied Basic Research (2023B0303000010).

\bibliographystyle{named}
\bibliography{ijcai25}

\end{document}